\documentclass[lettersize,journal]{IEEEtran}
\usepackage{amsmath,amsfonts}
\usepackage{algorithmic}
\usepackage{array}
\usepackage[caption=false,font=normalsize,labelfont=sf,textfont=sf]{subfig}
\usepackage{textcomp}
\usepackage{stfloats}
\usepackage{url}
\usepackage{verbatim}
\usepackage{graphicx}
\usepackage{booktabs}
\usepackage[figuresright]{rotating}

\def\BibTeX{{\rm B\kern-.05em{\sc i\kern-.025em b}\kern-.08em
    T\kern-.1667em\lower.7ex\hbox{E}\kern-.125emX}}
\usepackage{balance}
\usepackage{tabularx}
\begin{document}
\title{A Comprehensive Review of Multimodal Large Language Models: Performance and Challenges Across Different Tasks}
\author{Jiaqi Wang\textsuperscript{*}, Hanqi Jiang\textsuperscript{*}, Yiheng Liu\textsuperscript{*}, Chong Ma\textsuperscript{*}, Xu Zhang\textsuperscript{*}, Yi Pan\textsuperscript{*}, Mengyuan Liu\textsuperscript{*}, Peiran Gu\textsuperscript{*}, Sichen Xia\textsuperscript{*}, Lifeng Chen, Junhao Chen, Wenjun Li, Yutong Zhang, Zihao Wu, Zhengliang Liu, Tianyang Zhong, Bao Ge, Tuo Zhang, Ning Qiang, Xintao Hu, Xi Jiang, Xin Zhang, Wei Zhang, Dinggang Shen, Tianming Liu$\dagger$, Shu Zhang$\dagger$

\thanks{\textsuperscript{*}Major authors: Jiaqi Wang, Hanqi Jiang, Yiheng Liu, Chong Ma, Xu Zhang, Yi Pan, Mengyuan Liu, Peiran Gu, Sichen Xia}
\thanks{$\dagger$Corresponding authors: Tianming Liu, Shu Zhang.}
\thanks{Jiaqi Wang and Shu Zhang are with the School of Computer Science, Northwestern Polytechnical University, Xi'an 710072, China. Yiheng Liu, Chong Ma, Sichen Xia, Tianyang Zhong, Xintao Hu, and Tuo Zhang are with the School of Automation, Northwestern Polytechnical University, Xi'an 710072, China. Yutong Zhang, and Xin Zhang are with the Institute of Medical Research, Northwestern Polytechnical University, Xi'an, 710072, China. (e-mail: \{jiaqi.wang, liuyiheng123, mc-npu, sichenxia, wenjunlee, zyt20000316, 2022100670\}@mail.nwpu.edu.cn; \{tuozhang, xhu, xzhang, shu.zhang\}@nwpu.edu.cn)}

\thanks{Hanqi Jiang, Yi Pan, Lifeng Chen, Junhao Chen, Zihao Wu, Zhengliang Liu, and Tianming Liu are with the School of Computing, The University of Georgia, Athens 30602, USA. (e-mail: \{hj67104, ypan24, zihao.wu1, zl18864, tliu\}@uga.edu.)}

\thanks{Xu Zhang, Mengyuan Liu, Peiran Gu, Bao Ge, and Ning Qiang are with the School of Physics and Information Technology, Shaanxi Normal University, Xi’an 710119 China.  (e-mail: 2488009484@qq.com \{liumengyuan, \,bob\_ge, qn315\}@snnu.edu.cn, gupeiran123@gmail.com)}

\thanks{Xi Jiang is with the Clinical Hospital of Chengdu Brain Science Institute, MOE Key Lab for Neuroinformation, School of Life Science and Technology, University of Electronic Science and Technology of China, Chengdu, China (email: xijiang@uestc.edu.cn).}

\thanks{Wei Zhang is with the School of Computer and Cyber Sciences, Augusta University, Augusta, 30912, USA. (email: wzhang2@augusta.edu)}

\thanks{Dinggang Shen is with the School of Biomedical Engineering, ShanghaiTech University, Shanghai 201210, China; Shanghai United Imaging Intelligence Co., Ltd., Shanghai 200230, China; Shanghai Clinical Research and Trial Center, Shanghai, 201210, China. (e-mail: Dinggang.Shen@gmail.com).}

}

\markboth{Journal of \LaTeX\ Class Files,~Vol.~18, No.~9, September~2020}%
{How to Use the IEEEtran \LaTeX \ Templates}

\maketitle

\begin{abstract}
In an era defined by the explosive growth of data and rapid technological advancements, Multimodal Large Language Models (MLLMs) stand at the forefront of artificial intelligence (AI) systems. Designed to seamlessly integrate diverse data types—including text, images, videos, audio, and physiological sequences—MLLMs address the complexities of real-world applications far beyond the capabilities of single-modality systems. In this paper, we systematically sort out the applications of MLLM in multimodal tasks such as natural language, vision, and audio. We also provide a comparative analysis of the focus of different MLLMs in the tasks, and provide insights into the shortcomings of current MLLMs, and suggest potential directions for future research. Through these discussions, this paper hopes to provide valuable insights for the further development and application of MLLM.
\end{abstract}


\section{Introduction}
 
\IEEEPARstart{M}{LLMs} are sophisticated artificial intelligence (AI) systems designed to process and integrate various types of data, including text, images, videos, audio, and physiological sequential data~\cite{wu2023multimodal,yin2023survey,zhang2024mm}. As we navigate the era of multimodal data fusion, marked by rapid advancements in information technology and an explosive increase in data volume, the capabilities of single-modality systems no longer suffice for complex real-world tasks~\cite{koh2024generating,ma2024eye,zhang2024potential}. Thus, the development of MLLMs is not only an inevitable trend in technological evolution but also a critical enhancement for improving the effectiveness of AI applications. By synergizing information from multiple data sources, MLLMs cultivate a more comprehensive and accurate representation of information, this capability not only unlocks significant potential but also demonstrates substantial practical application value across a diverse range of fields. The integration of diverse datasets tailors MLLMs to perform more effectively, establishing them as indispensable next-generation techniques in the ongoing quest to harness the full potential of AI technologies~\cite{zhang2023biomedgpt,mei2024phraseaug,xiao2024instruction}. Notably, MLLMs have shown remarkable performance across diverse multimodal tasks, including language, image, video, and audio processing. These models excel in integrating multimoda information to enhance the effectiveness of multimodal tasks. 

In Natural Language Processing (NLP) tasks such as text generation and machine translation, MLLMs leverage images, video, and audio to provide contextual support, enhancing the accuracy and expressiveness of the generated text~\cite{malik2021automatic,bahar2019comparative,lyu2023macaw}. These models also excel in sentiment analysis and dialog systems by integrating multimodal information to improve understanding and generation capabilities. In particular, MLLMs transform NLP by incorporating visual and auditory data, thus enriching text generation and machine translation~\cite{achiam2023gpt,zheng2024judging,le2023bloom}. These models enhance the accuracy and expressiveness of the generated text, providing nuanced contextual support that traditional models cannot. In sentiment analysis and dialogue systems, MLLMs are capable of the integration of multimodal information, which further deepens the understanding of system and response capabilities, presenting a leap forward in human-computer interaction~\cite{wang2024largelanguagemodelsrobotics,wang2023review}.

In addition, in the vision tasks, MLLMs significantly enhance improves task comprehension, analysis, and generation. Integrating textual descriptions and image instructions commands allows for more accurate tasks such as image classification, target detection, and image annotation. For instance, MLLMs such as GPT-4V~\cite{achiam2023gpt}, and Gemini~\cite{team2023gemini} combine image content with natural language descriptions to produce more vivid and precise annotation results. These models also show progress in image generation, creating images from textual descriptions or enabling cross-modal image style migration, thus broadening the possibilities within this field. Meanwhile, video processing poses unique challenges due to its complexity. Nevertheless, the advent of MLLMs has propelled forward the capabilities of language models in this domain. Models like NExT-GPT~\cite{wu2023next} and Sora~\cite{liu2024sora} are pioneering multimodal video generation, producing richer and more realistic video content by learning from multimodal data. Furthermore, advancements in intelligent video understanding technologies, such as VideoChat~\cite{li2023videochat} and Video-LLaVA~\cite{lin2023video}, have significantly enhanced the ability to analyze and process video content. These developments promise enhanced user experiences in virtual reality, video games, and educational applications.

Furthermore, In audio tasks, MLLMs bring a new technological change to audio processing tasks. Traditional audio processing usually relies on unimodal signal processing methods, such as speech recognition~\cite{gaikwad2010review} or audio classification~\cite{mulimani2024class}, which have limitations in processing complex multimodal data.MLLMs are able to understand better and generate audio-related content by combining audio signals with textual and visual information through the combination of Large Language Models (LLMs). For example, in speech generation tasks, MLLMs can utilize textual and visual information to generate more natural and contextually relevant speech output~\cite{shen2024hugginggpt,huang2024audiogpt}. In audio understanding tasks, these models can perform sentiment recognition, audio classification, or audio event detection more accurately by combining visual cues and textual descriptions. In addition, MLLMs show strong potential in tasks such as cross-modal audio-text translation, audio soundtrack generation, and multimodal sentiment analysis~\cite{tang2023salmonn,team2023gemini}. These technological advances not only improve the effectiveness of audio processing, but also expand its scenarios in real-world applications such as smart homes, virtual assistants, movie and television production, etc.

\begin{figure*}[ht]
\begin{center}
\includegraphics[width=0.9\textwidth]{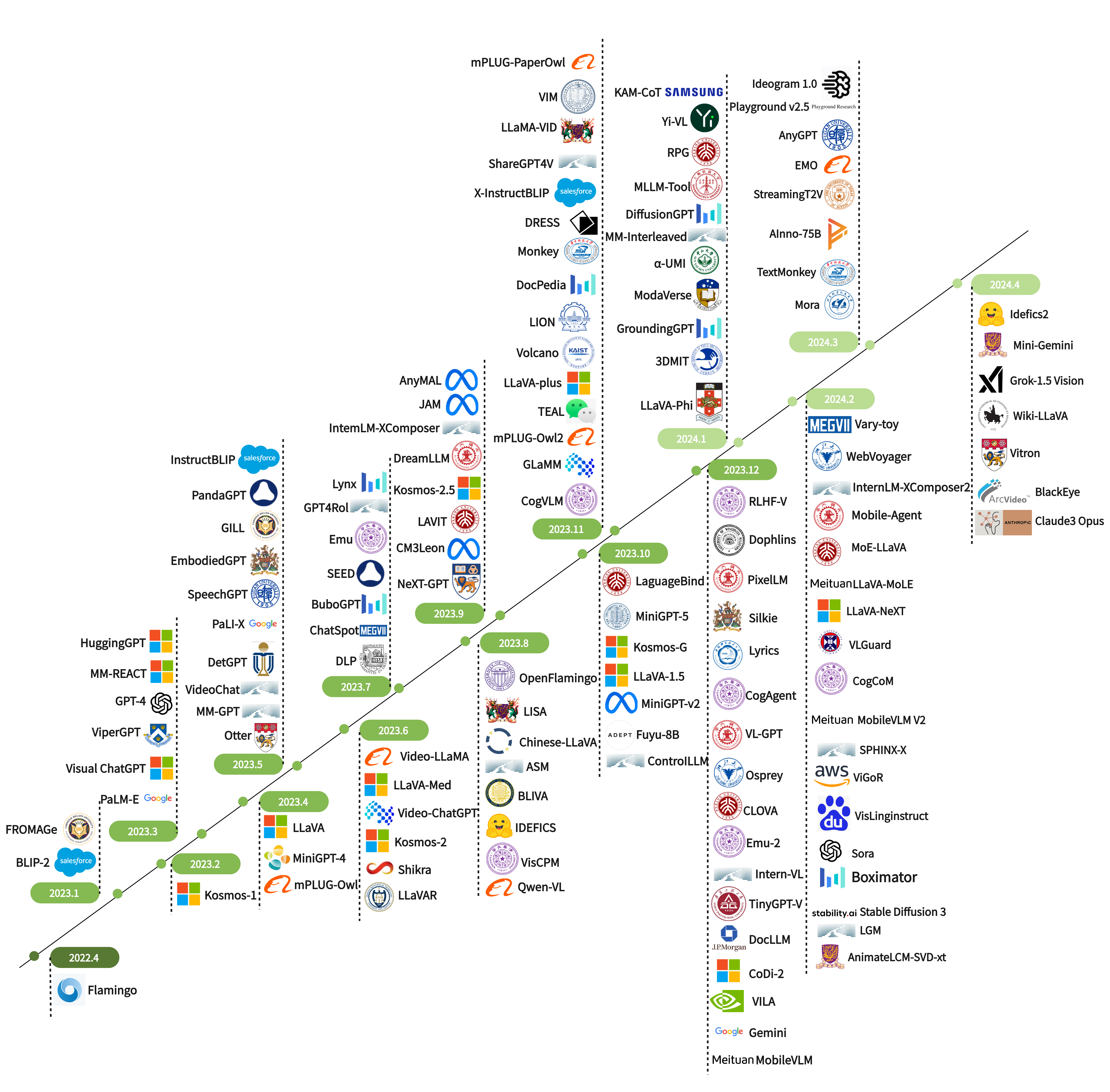}

\caption{A timeline of representative MLLMs.
} 
\end{center}
\label{Fignure}
\end{figure*}

As shown in Fig~\ref{Fignure}, we summarize MLLMs from recent years. This paper reviews the current state of the art of MLLM applications, introduces the basic concepts and main architectures of MLLMs in Section 2, describes their performance in different domains to identify their strengths and weaknesses in Section 3, highlights the transformative impact of MLLMs through a comparative analysis in Section 4, and provides a roadmap for future research in Section 5. Our discussion aims to incentivize continuous innovation and ensure that MLLMs remain at the forefront of AI technology development. Through a comprehensive review of current implementations and progress, this paper aims to summarize research results, provide valuable references, and offer guidance for future research in the field of MLLMs. Our goal is to inspire new ideas and directions for the continued development of MLLMs, ensuring that they remain at the forefront of AI technology development.


\section{Overview of Multimodal Large Language Models}
\subsection{Definitions and Basic Concepts}
Overall, MLLMs represent a significant advancement in the field of artificial intelligence and machine learning, embodying the capability to process and interpret a diverse data type including text, images, audio, and video \cite{li2024multimodal,liu2023summary,liu2024understanding}. By integrating and synthesizing data from these different modalities, MLLMs achieve a more comprehensive and precise understanding and generation of information \cite{zhang2024mm}.

In particular, MLLMs are sophisticated and comprehensive systems engineered to handle and decode multimodal data concurrently. The core principle of MLLMs lies in the integration and interplay of varied modalities, which significantly amplifies the effectiveness of the models. This multimodal approach not only enhances the understanding of individual data types but also fosters a more nuanced interaction between them, thereby expanding the scope and accuracy of AI applications. For instance, in tasks such as image captioning, MLLMs leverage both text and visual data to produce accurate and contextually relevant descriptions of images. This synergy enables the models to surpass the limitations of single-modality systems, offering richer and more detailed outputs. Additionally, the combination of audio and visual data can greatly improve the performance of tasks like video understanding and annotation, making MLLMs invaluable for applications requiring detailed multimedia analysis.

By harnessing the collective strengths of various data types, MLLMs not only enhance the capability of AI to interpret and interact with the world but also pave the way for groundbreaking developments in how machines understand complex, multifaceted information.

\subsection{Main Components of Multimodal Large Language Models}
A MLLM harnesses several vital components to efficiently process and integrate data from diverse modalities. These components are designed to transform raw input from various sources into actionable insights, tailoring these models incredibly versatile and effective. The architecture of these models can be broadly categorized into three main components: the multimodal input encoder, the feature fusion mechanism, and the multimodal output decoder.

\subsubsection{Mutimodal Input Encoder} The multimodal input encoder is a vital component in MLLMs, designed to transform raw input data from various modalities into a structured format that the model can process effectively. This crucial module is specialized in handling different types of data, ensuring that each data form is optimally encoded to contribute effectively to the model's overall function. Here's how the encoder works for each data type: 

\textbf{Text:} For textual data, the encoder utilizes technologies like embedding layers, which map words into vectors of continuous numbers, and Multi-Layer Perceptrons (MLP), or more advanced transformers that manage long-range dependencies and context within text. 

\textbf{Images:} Visual data is processed using state-of-the-art architectures such as Vision Transformers (ViT) \cite{dosovitskiy2020image}, which treat parts of an image as sequences to better capture relationships, or Residual Networks (ResNet) \cite{he2016deep}, which help in learning deeper features without losing context through layers. 

\textbf{Audio:} Audio data is analyzed using models like C-Former \cite{chen2023x}, HuBERT \cite{hsu2021hubert}, BEATs \cite{chen2022beats}, or Whisper \cite{radford2023robust}. Each of these models is tailored to capture the unique properties of sound, from basic tones to complex spoken language, enhancing the model’s ability to interpret auditory information accurately. 

\textbf{Sequential Data:} For sequential data such as EEG and heartbeats, the encoder employs a combination of 1D-Convolutional Neural Networks (1D-CNN) and Long Short-Term Memory (LSTM) units. This setup is particularly effective in picking up temporal and spatial patterns in data, which is crucial for early diagnostics in medical applications. 

\textbf{Universal Encoder:} A more recent innovation is the universal encoder, which aims to standardize the encoding process across highly diverse data types, including audio, video, and functional Magnetic Resonance Imaging (fMRI). This encoder leverages a generalized approach to handle and integrate multiple forms of data, promoting consistency and efficiency in data processing. Each of these encoders converts raw inputs into feature vectors, which are then transformed into a fixed-length feature sequence. This standardization is critical as it prepares the data for further processing, ensuring that the subsequent layers of the model can perform feature fusion and decoding effectively.

By accommodating and optimizing the initial processing of varied data types, the multimodal input encoder not only enhances the model’s performance but also expands its applicability across different fields. Whether it's improving the accuracy of image captioning, enriching the context of machine translation, or advancing the precision of diagnostic tools in healthcare, this encoder plays a foundational role in enabling AI models to perform complex tasks that require a nuanced understanding of diverse inputs.

\subsubsection{Feature Fusion Mechanism} The heart of a multimodal model is its ability to integrate features from different modalities. This integration can occur at various stages~\cite{gadzicki2020early,li2024review}: 

\textbf{Early Fusion:} Combines input data at the initial stage, leveraging the raw interconnectedness of different modalities. 

\textbf{Intermediate Fusion:} Merges features during the feature extraction phase, allowing each modality to contribute its unique properties to a unified representation. 

\textbf{Late Fusion:} Integrates the final outputs from individual modal pathways at the decision stage, often used in tasks requiring consolidated judgments from multiple data types. 

\textbf{Joint Fusion:} A hybrid approach that merges early, intermediate, and late fusions to maximize data utilization across all stages. These fusion processes often employ pre-trained LLMs, which, while initially designed for textual data, are adapted to handle and synthesize multimodal inputs through advanced feature projection and serialization techniques.

\textbf{Multimodal Output Decoder:}  Lastly, the multimodal output decoder reconverts the fused, integrated multimodal information back into a usable form tailored to specific tasks, such as \textbf{Image captioning}, the decoder might generate descriptive text based on visual inputs. \textbf{Video understanding tasks}, it could produce annotations or summaries combining both visual and auditory data. Each decoder is meticulously designed to optimize accuracy and quality, ensuring that the output precisely reflects the combined insights gained from the integrated modalities.

To summarize, the sophisticated architecture of multimodal large language models empowers them to tackle complex tasks by harnessing and synthesizing data across text, images, and audio. This capability not only enhances the performance of AI applications but also opens up new avenues for innovation in how we understand and interact with technology.

\subsection{Overview of Multimodal Feature in LLMs}
When fusing multimodal features, it is common practice not to train new models from scratch. Instead, existing pre-trained large models, such as LLMs, are utilized. Although pre-trained LLMs are primarily designed to handle text input, various techniques can be employed to adapt these models for processing multimodal data. We will introduce a specific example in this section to illustrate the fusion process and understand it in detail.

First, data from each modality needs to be encoded and projected into a unified feature space. For instance, pre-trained models like ResNet or Vision Transformer can be used to convert image data into feature vectors $V_{image}$. Text data can be converted into feature vectors $V_{text}$ using pre-trained text encoders like BERT~\cite{kenton2019bert}, and audio data can be transformed into feature vectors $V_{audio}$ using pre-trained audio encoders like wav2vec~\cite{schneider2019wav2vec}. Then, the feature vectors from different modalities are mapped into a shared feature space through linear transformations or other projection methods.To input these multimodal features into a pre-trained LLM, the features from different modalities need to be organized into a sequence. A multimodal feature sequence can be formed by simply concatenating the features from different modalities, such as $[V_{image}, V_{text}, ..., V_{audio}, V_{text}]$. 

Next, the constructed multimodal feature sequence is fed into the pre-trained LLM for processing. The Transformer model processes the input feature sequence through multiple layers of self-attention mechanisms and feedforward neural networks. Each layer, comprising self-attention and feedforward network modules, updates and integrates the feature representations, progressively extracting higher-level features. After passing through multiple Transformer layers, the model generates a feature representation sequence that encapsulates comprehensive information. Depending on the task requirements, a specific output layer can generate the final result. For instance, if the task is to generate a textual description, the integrated feature representation can be fed into a text generator to produce descriptive text.

By following these steps, multimodal features can be effectively processed by an LLM. Although pre-trained language models like GPT, LLAMA are primarily designed for text input, their capabilities can be extended to handle and integrate multimodal data through feature projection and serialization methods, enabling them to perform complex multimodal tasks.

\section{Task Classification of Multimodal Large Language Models}
In this section, we will detail the contribution of MLLMs in visual and audio tasks in terms of its application to specific tasks. At the same time, we will explore how to construct MLLMs suitable for these task types.

\subsection{Image Tasks}

MLLMs represent a significant advancement in AI industry by integrating visual and textual data to perform complex tasks. These models are designed to understand and generate images, leveraging their ability to process and combine different types of information. This section explores the two primary tasks of MLLMs: image understanding and image generation, illustrating their capabilities and applications.

In the realm of image understanding, MLLMs excel at interpreting and extracting meaningful information from images. They can identify objects, understand scenes, and even infer relationships between elements within an image. By combining visual data with textual descriptions, these models can provide comprehensive analyses that surpass traditional image processing methods. For instance, an MLLM can analyze a photograph to identify the objects present, describe their interactions, and correlate this with relevant textual information, offering a nuanced understanding of the visual content.

On the other hand, the image generation capabilities of MLLMs are equally impressive. These models can create realistic images from textual descriptions, generate variations of existing images, and even produce entirely new visual content that aligns with given textual inputs. This generative ability is crucial for applications such as creative industries, where generating novel visual content is essential. By training on vast datasets of images and texts, MLLMs can synthesize new images that are coherent with the specified descriptions, pushing the boundaries of what automated image creation can achieve.

The integration of image understanding and generation tasks in MLLMs highlights their versatility and potential. By effectively bridging the gap between visual and textual modalities, these models open up new possibilities for advanced AI applications. They enable more intuitive human-computer interactions, where users can receive detailed visual analyses or generate desired images through simple textual commands. This dual capability not only enhances the user experience but also expands the potential use cases across various fields, from automated content creation to enhanced data interpretation.

In the following contents, the image tasks of MLLMs encompass two main areas: image understanding and image generation. These capabilities showcase the models' ability to analyze and create visual content by integrating and leveraging multimodal data. The advancements in these tasks underscore the transformative impact of MLLMs in the realm of AI, promising continued innovation and expanded applications.

\subsubsection{Image Understanding}
\ 
\newline
\textbf{Task Description}: 

The development of image understanding techniques in MLLMs has gone through several stages, from the earliest reliance on traditional feature extraction methods to the current widespread application of deep learning technologies, resulting in a diverse range of research achievements and innovations. The following are the developmental stages of image understanding techniques in MLLMs: 1. Image understanding based on traditional feature extraction methods; 2. Application of deep learning technologies in image understanding; 3. Multimodal image understanding and cross-modal learning; 4. Application of reinforcement learning in image understanding; 5. Integration of image generation and understanding.

\paragraph{Image Understanding Based on Traditional Feature Extraction Methods}

In the early days, image understanding primarily relied on traditional feature extraction methods such as HOG and SIFT. These methods involved manually designing features to describe image content, followed by utilizing traditional machine learning algorithms for tasks like classification and detection. While these methods performed well in some simple image tasks, they exhibited limitations in descriptive capabilities and generalization when faced with complex image data.

\paragraph{Application of Deep Learning Technologies in Image Understanding}

With the rapid advancement of deep learning technologies, image understanding entered a new phase. Through techniques like deep convolutional neural networks (CNN), it became possible to learn high-level feature representations directly from raw pixel data, enabling more precise image classification, object detection, image segmentation, and other tasks. CNN networks achieved breakthroughs in image understanding, as evidenced by their victory in the ImageNet image classification competition, marking the widespread application of deep learning technologies in the field of image understanding.

\paragraph{Multimodal Image Understanding and Cross-Modal Learning}

As NLP and computer vision intersect, multimodal image understanding has become a research hotspot. By integrating multimodal data such as images and text, tasks like image description and visual question answering can be achieved. Additionally, the application of cross-modal learning technologies has opened up new possibilities for image understanding by fusing different modal information to enhance the comprehension and representation of image content.

\paragraph{Application of Reinforcement Learning in Image Understanding}

In recent years, the application of reinforcement learning in image understanding has been increasing. Reinforcement learning can help models make decisions in complex environments, thereby enhancing their ability in image understanding. When combined with deep learning technologies, reinforcement learning can achieve more intelligent and adaptive image understanding processes in tasks such as image classification and object detection.

\paragraph{Integration of Image Generation and Understanding}

Recent research indicates that combining image generation and understanding can yield more comprehensive and in-depth information comprehension. Through generative models like Generative Adversarial Networks (GAN), images with semantic flexibility can be generated to assist models in better understanding image content. This integration brings more inspiration and innovation to image understanding tasks, while also providing additional support and application scenarios for image generation technologies.

In conclusion, the image understanding techniques in s have undergone several developmental stages, including traditional feature extraction, deep learning, multimodal learning, reinforcement learning, and the integration of image generation and understanding. The continuous innovation and development of these techniques have propelled the progress of image understanding, bringing forth more possibilities and opportunities for research and application in the field of computer vision. As artificial intelligence technology continues to evolve, the future development of image understanding techniques will become more diverse and intelligent, providing stronger support for achieving more intelligent image understanding and applications.

\ 
\newline
\textbf{Model Introduction}: 

Multimodal models play a crucial role in image understanding tasks by integrating text and image information to achieve more intelligent and comprehensive understanding and reasoning. In this field, MiniGPT-4~\cite{zhu2023minigpt,chen2023minigptv2} ,InstructBLIP~\cite{instructblip} ,and Wiki-LLaVA~\cite{caffagni2024wiki} are three highly regarded models, along with other related models such as 3DMIT~\cite{li20243dmit}, GroundingGPT~\cite{li2024groundinggpt}, ModaVerse~\cite{wang2024modaverse}, Vary-toy~\cite{wei2024small}, LLaVA-MOLE~\cite{chen2024llava}, and CogCom~\cite{qi2024cogcom}.

MiniGPT-4 is a multimodal model based on GPT-4 that combines text and image information, exhibiting excellent image understanding capabilities. MiniGPT-4 adopts a modular architecture design, which allows for rapid adaptation to various LLMs, enabling flexible understanding and reasoning for different types of images. The model utilizes advanced image encoders and text generators to accurately describe and infer information from images, providing strong support for image understanding tasks.

Another prominent model is InstructBLIP, which demonstrates exceptional image understanding capabilities. With its modular architecture design, InstructBLIP can quickly adapt to different LLMs and perform instruction tuning by freezing the image encoder and the LLM. This approach enables efficient understanding and reasoning of image information. InstructBLIP excels in vision-language instruction tuning, generating rich responses while effectively considering the dialogue history, thus providing impressive performance for image understanding tasks.

Besides MiniGPT-4 and InstructBLIP, there are other noteworthy multimodal models that also play significant roles in image understanding tasks. 3DMIT employs advanced three-dimensional information processing technology, enabling deep understanding and analysis of 3D images. GroundingGPT focuses on effectively correlating natural language and visual information to achieve cross-modal understanding and reasoning. Models like ModaVerse, Vary-toy, LLaVA-MOLE, and CogCom also exhibit outstanding image understanding capabilities to varying degrees, each with unique features and advantages, contributing significantly to image understanding tasks.

In summary, multimodal models play an essential role in image understanding tasks. MiniGPT-4, InstructBLIP, and other related models have shown excellent performance and potential, providing strong support for the development of image understanding tasks. With continuous technological advancements and improvements in model performance, multimodal models are expected to play an increasingly important role in image understanding tasks, bringing more innovation and breakthroughs to the field of artificial intelligence.

\textbf{MiniGPT-4}

\textbf{Architecture:}
MiniGPT-4 is designed to align visual information from a pre-trained vision encoder with an advanced LLM. Specifically, it utilizes Vicuna as the language decoder, which is built upon LLaMA and can perform a wide range of complex linguistic tasks. For visual perception, the model employs the same visual encoder used in BLIP-2, consisting of a ViT backbone coupled with a pre-trained Q-Former. Both the language and vision models are open-sourced. A linear projection layer bridges the gap between the visual encoder and the LLM. The architecture allows MiniGPT-4 to handle diverse vision-language tasks without relying on external vision models.

\textbf{Datasets:}
During the first pre-training stage, MiniGPT-4 is trained using a large collection of aligned image-text pairs. This includes datasets like Conceptual Caption, SBU, and LAION, amounting to approximately 5 million image-text pairs. In the second fine-tuning stage, a smaller yet high-quality dataset is curated, which includes detailed image descriptions tailored for vision-language alignment. This stage involves only 3,500 high-quality image-text pairs designed in a conversational template to enhance the model's generation reliability and usability.

\textbf{Training \& Evaluation:}
MiniGPT-4 employs a two-stage training approach. The initial pre-training stage involves training the model on a large collection of aligned image-text pairs, keeping the pre-trained vision encoder and the LLM frozen while only training the linear projection layer. This stage spans 20,000 training steps with a batch size of 256, taking approximately 10 hours on 4 A100 GPUs. The second fine-tuning stage addresses the limitations observed after the first stage, such as incoherent linguistic outputs. It fine-tunes the model using a smaller but high-quality dataset, significantly improving its generation reliability. This fine-tuning is computationally efficient, taking only around 7 minutes with a single A100 GPU.

\textbf{InstructBLIP}

\textbf{Architecture:}
InstructBLIP builds upon the BLIP-2 architecture by incorporating an instruction-aware Q-Former module. This module takes in instruction text tokens as additional input, interacting with the query embeddings through self-attention layers. This process encourages the extraction of task-relevant image features, allowing the LLM to receive visual information conducive to following instructions. InstructBLIP uses the same image encoder as BLIP-2, a ViT backbone, and adapts the model to different frozen LLMs, including FlanT5-XL, FlanT5-XXL, Vicuna-7B, and Vicuna-13B. The architecture facilitates instruction-aware visual feature extraction, substantially improving performance in both held-in and held-out evaluations.

\textbf{Datasets:}
InstructBLIP is trained using a comprehensive set of 26 publicly available datasets, transformed into an instruction tuning format. These datasets cover a wide range of tasks such as image captioning, visual question answering (VQA), and visual instruction following. Notable datasets include OK-VQA, ScienceQA, HatefulMemes, Visual Dialog, MSVD, and MSRVTT. The training process balances the datasets to prevent overfitting smaller datasets and underfitting larger ones, ensuring effective learning across varied tasks.

\textbf{Training \& Evaluation:}
InstructBLIP employs a two-stage training process. The first stage involves vision-language pre-training, initializing the model from pre-trained BLIP-2 checkpoints. Only the parameters of the Q-Former are fine-tuned while keeping the image encoder and the LLM frozen. The second stage, instruction tuning, involves training the model with a maximum of 60,000 steps, validating performance every 3,000 steps. The AdamW optimizer is used with specific hyperparameters, and training is conducted on 16 Nvidia A100 GPUs over 1.5 days. Inference involves generating responses for image captioning and open-ended VQA tasks, and using vocabulary ranking for classification and multi-choice VQA tasks.

\subsubsection{Image Generation}
\ 
\newline
\textbf{Task Description}: 

The application of multimodal models in image generation tasks is one of the hot research directions in the field of computer vision. Traditional image generation tasks mainly rely on single-modal data input, such as generating images from random noise or generating images based on context. In contrast, multimodal models simultaneously consider multiple data modalities, such as text descriptions, speech information, etc., to achieve more accurate and diversified image generation.

In image generation tasks, the application of multimodal models is mainly reflected in the following aspects: First, by integrating information from different modalities, multimodal models can achieve conditional image generation tasks. For example, given a text description, multimodal models can learn the correspondence between text and images, thus generating images that match the description. This approach makes image generation more targeted and semantically coherent, improving the quality and accuracy of generated images.

Secondly, the application of multimodal models in image generation also manifests in cross-modal generation tasks. For example, applying artistic styles from one image to another to achieve image style transfer; or generating corresponding speech descriptions based on image content. These tasks require the model to effectively learn the correlated information between different modal data, thereby realizing the transformation and generation of cross-modal information, bringing more creativity and imagination to image generation tasks.

The technological development of image data generation in s can be roughly divided into the following stages: 1. Image generation based on Generative Adversarial Networks (GAN), 2. Improvement and optimization of image generation models, 3. Multimodal generation combining images and text, 4. Application of transfer learning and self-supervised learning in image generation.

\paragraph{Early Stage of Image Data Generation in s based on GAN}

GAN as a deep learning architecture were widely used for generating realistic images in the early stages of image data generation in s. GAN consists of a generator and a discriminator, and it improves the capability of the generator to produce realistic images through adversarial training. Despite the significant success of GAN in image generation, its training process faced challenges such as mode collapse and training instability. Researchers continuously optimized GAN models and introduced new techniques and methods during the training process to address these issues.

\paragraph{Improvement and Optimization of Image Generation Models}

To address the challenges of GAN, researchers proposed various improvement and optimization techniques. For example, the introduction of conditional Generative Adversarial Networks (cGAN) achieved image generation under specific conditions; Wasserstein GAN (WGAN) improved training stability by enhancing the loss function; upgraded versions of Generative Adversarial Networks (GAN) such as ProGAN, StyleGAN, etc., could generate higher resolution and more realistic images. These improvement and optimization measures significantly enhanced the quality and stability of image generation models.

\paragraph{Multimodal Generation combining Images and Text}

With the fusion development of NLP and computer vision, researchers began studying multimodal generation combining images and text. This phase focused on integrating text descriptions with image generation, achieving tasks such as image description generation, visual question answering, and more. Multimodal generation technology brought wider applications and innovations to image generation tasks, expanding the diversity and scope of image generation.

\paragraph{Application of Transfer Learning and Self-Supervised Learning in Image Generation}

In recent years, techniques such as transfer learning and self-supervised learning have been widely applied in image generation tasks. Transfer learning effectively transfers learned knowledge to new tasks, improving the effectiveness of image generation, while self-supervised learning utilizes intrinsic label information from data to learn effective representations, enhancing the performance of generative models. The application of these technologies brings more possibilities and efficiency improvements to image generation tasks, driving the continuous advancement of image data generation in s.

In conclusion, the task of image generation in MLLMs represents a dynamic and rapidly evolving area of computer vision. By leveraging the integration of multiple data modalities, these models have significantly enhanced the precision, diversity, and creativity of image generation. The progression from traditional GAN-based methods to more sophisticated techniques incorporating transfer learning and self-supervised learning underscores the continuous innovation driving this field. Multimodal models enable conditional and cross-modal image generation, providing semantically coherent and targeted outputs that open new avenues for applications in various domains. As these technologies advance, they promise to further refine and expand the capabilities of image generation, solidifying the role of MLLMs at the forefront of artificial intelligence research.

\ 
\newline
\textbf{Model Introduction}: 

In current multimodal deep learning research, image generation tasks have garnered significant attention, leading to the emergence of several innovative models. This article introduces several representative models, showcasing their applications and innovations in image generation tasks.

Firstly, let's consider ProGAN~\cite{gao2019progan}. ProGAN builds upon the concept of progressive growing for generative adversarial networks. ProGAN incrementally increases the resolution of generated images during the training process, starting from low-resolution images and gradually adding layers to enhance details and complexity. This approach significantly stabilizes the training dynamics and improves the quality of the generated images, reducing artifacts that were common in earlier GAN models. ProGAN's ability to produce high-definition images with resolutions up to 1024x1024 pixels has demonstrated remarkable performance across various benchmarks, making it a pivotal advancement in the field of image generation.

Another innovative model is MM-Interleaved~\cite{tian2024mm}, which builds upon the Multimodal Feature Synchronizer (MMFS). MMFS improves efficiency in extracting visual details for MLLMs by reducing the required number of visual annotations. MM-Interleaved optimizes the interaction between text and images through end-to-end training, particularly suitable for tasks requiring dynamic and efficient image attention mechanisms. It demonstrates remarkable performance across various multimodal benchmarks, highlighting its potential and advantages in image generation tasks.

In addition to these mainstream models, there are notable models focused on specific domains and tasks. For instance, AnimateLCM~\cite{wang2024animatelcm} utilizes Singular Value Decomposition (SVD) and Long Short-Term Memory (LSTM) networks for animation generation, showcasing unique technical strengths in handling sequential data and generating dynamic images. Models like Ideogram 1.0~\cite{ideogram}, and Playground v2.5~\cite{li2024playground} excel in joint text-image generation with distinct features: TextMonkey uses reinforcement learning and Variational Autoencoders (VAE) for optimized generation; Ideogram 1.0 specializes in generating symbols and graphics, widely applicable in data visualization and information graphics; Playground v2.5 serves as an open platform exploring diverse multimodal generation implementations, offering researchers space for testing and optimizing models.

Furthermore, models like Idefics2~\cite{laurenccon2024matters}, and Mini-Gemini~\cite{li2024mini} each emphasize complex data generation, resource efficiency, and semantic information integration with visual expression, respectively. Idefics2 combines semantic and visual features for image generation in cross-modal tasks; Mini-Gemini excels as a lightweight model advantageous in resource-constrained or fast inference scenarios; Wiki-LLaVA emphasizes the integration of semantic information and visual expression, particularly suited for image-text scenarios.

In summary, these multimodal generation models illustrate diverse and innovative development paths—from integrating language models to optimizing visual attention mechanisms, to focusing on specific tasks such as animation or symbol graphic generation. They not only enhance generation quality but also expand the practical applications of multimodal deep learning. With further advancements in algorithms and hardware, these models are poised to exhibit broader application prospects and higher performance levels across various domains in the future.

\textbf{ProGAN}

\textbf{Architecture}: ProGAN, or Progressive Growing of GANs, employs a unique architecture where both the generator and discriminator grow progressively during training. This means starting from a low resolution, new layers are added incrementally to model increasingly fine details as training progresses. The generator and discriminator both start with a 4x4 pixel resolution and grow to a maximum resolution, such as 1024x1024 pixels, by doubling the resolution at each step. This incremental approach stabilizes training and improves image quality by allowing the model to learn high-level structures before focusing on fine details. Key innovations include pixel-wise feature vector normalization in the generator and minibatch standard deviation in the discriminator, which help maintain the stability and variation of the generated images throughout the training process.

\textbf{Datasets}: ProGAN is primarily trained on high-quality datasets such as CelebA-HQ, which consists of high-resolution images of celebrities. This dataset is essential for the model to learn to generate photorealistic human faces. Additionally, ProGAN has been tested on other large-scale datasets like LSUN, which includes categories such as churches, bedrooms, and outdoor scenes, allowing the model to generate a diverse range of photorealistic images beyond just human faces.

\textbf{Training \& Evaluation}: The training of ProGAN involves a novel technique where the generator and discriminator are progressively grown. This method starts with a simple 4x4 resolution and progressively adds layers to increase the resolution of the generated images. The training uses the Wasserstein GAN loss with a gradient penalty (WGAN-GP) to improve stability and convergence. Training is performed on multiple GPUs, such as 8 Tesla V100 GPUs, over several days to ensure high-quality results. Evaluation of ProGAN shows its capability to generate highly realistic images with minimal artifacts, as demonstrated on benchmark datasets like CelebA-HQ and LSUN.

\textbf{MM-Interleaved}:

\textbf{Architectures:}
MM-Interleaved is an end-to-end generative model designed for interleaved image-text data. It integrates a Visual Foundation Model (VFM), a LLM, and a Diffusion Model (DM) to handle both text and image generation tasks effectively. The architecture comprises three key components: a VFM-based Image Tokenizer that uses a pre-trained vision model, such as CLIP-ViT, to extract image features, which are then processed by a Perceiver Resampler to map each image to a fixed number of visual tokens; an LLM-based Multi-modal Model that utilizes a pre-trained LLM, such as Vicuna, to extract context features from interleaved image-text sequences; and a Diffusion Model used for generating images from text inputs by leveraging the strengths of denoising-diffusion processes. The architecture is optimized end-to-end to preserve fine-grained image details and handle multiple images efficiently, reducing the computational and memory demands typically associated with multi-image scenarios.
\textbf{Datasets:}
The training of MM-Interleaved involves a two-stage approach. First, the model is pre-trained using a mixture of large-scale image-text datasets, including LAION-COCO and LAION-En, which provide extensive image-text pairs. The use of webdataset ensures efficient data loading and handling during the pre-training phase. For fine-tuning, MM-Interleaved employs datasets designed for complex multi-modal instructions, such as MMC4, OK-VQA, and VQA. These datasets help the model learn to follow complex multi-modal instructions and generate coherent text and image outputs based on interleaved sequences.
\textbf{Training \& Evaluation:}
MM-Interleaved's training process is structured in two stages. The first stage, single-modal pre-training, involves pre-training the modality-specific components (VFM and LLM) separately on large-scale image-text paired data to align their outputs with the LLM's word embedding space. The second stage, multi-modal instruction tuning, involves fine-tuning the integrated model using a high-quality multi-modal instruction-following dataset. This process includes both positive and negative examples to enhance the model's ability to handle arbitrary combinations of input modalities. The training utilizes DeepSpeed ZeRO-1 for efficient distributed training, ensuring scalability and performance during large-scale training sessions. Evaluation is conducted using zero-shot benchmarks across various datasets, ensuring the model's versatility and effectiveness in recognizing visual details and generating consistent outputs based on both textual and visual conditions.

\subsection{Video Tasks}
With the rapid development of information technology, video has become the primary means for people to access information and enjoy entertainment, playing an increasingly important role in today's social media and internet culture. 
However, manually processing such a vast amount of video content has proven to be a time-consuming and labor-intensive task. Effectively understanding and processing video content, especially when the volume of video data is large, the information is dynamic, and it is multimodal, has long been a major challenge in the technology field. In recent years, with the rise of MLLMs, the field of MLLMs has experienced rapid and flourishing development in videos domains. 
This section first introduces detailed explanations of two mainstream tasks in video fields: video generation and video understanding. Then, we introduce the development of multimodal large language models in these tasks. Finally, we detail the architecture, datasets, training methods, and evaluations of several representative models.

\subsubsection{Video Understanding Task}
\ 
\newline
\textbf{Task Description}: 

Current mainstream video understanding algorithms cover several foundational areas, such as action recognition, temporal action localization, video question answering (VQA), and video retrieval. Compared to pure text and images, videos have a multimodal nature, containing synchronized audio, text, and visual information, which provide rich data sources for video understanding analysis. 
Specifically, action recognition is primarily used to identify actions occurring in video segments, focusing on temporal continuity and development. Temporal action localization aims to locate the time intervals in long videos where specific actions occur, achieving seamless connections between visual content and textual context. Video question answering (VQA) requires providing correct answers to questions based on given video inputs, emphasizing the inference of logical implicit content from videos. This highlights the model's understanding of the visual and auditory components of the video, as well as the integration of external knowledge and reasoning abilities to provide contextually relevant answers. Video retrieval requires understanding both the language domain and the visual domain, and employing appropriate matching strategies to extract the most relevant videos from larger video repositories based on textual descriptions.

\noindent{\textbf{Model Introduction}:}

The surge in applications of s for video comprehension has led to significant advancements in video-related tasks, such as GPT-4~\cite{achiam2023gpt}, LLaMA~\cite{touvron2023llama}, Vicuna~\cite{zheng2024judging}, and BLOOM~\cite{le2023bloom}. These models rely on their powerful language understanding capabilities to follow instructions provided by humans and provide corresponding responses. The latest advancements in multimodal understanding are largely based on the combination of pre-trained image models and LLMs.

Due to the inherent feature space differences between pre-trained image models and LLMs, mainstream methods typically employ a unique projection layer to align the image features with the feature space of the LLM. Subsequently, features from multiple modalities are input into the LLM for further interaction. Based on the modality interaction architecture, these models can be categorized into two types. The first type relies entirely on the LLM's capability to handle multimodal feature interactions, while the second type performs preliminary multimodal feature interaction before the LLM.

Models using the first type of interaction include VILA~\cite{lin2024vila}, LLaVA-NeXT-Video\cite{zhang2024llavanext-video}, Mantis~\cite{jiang2024mantis}, Video-LLaMA~\cite{zhang2023video}, PG-Video-LLaVA~\cite{munasinghe2023pg}, Video-LLaMA-2~\cite{cheng2024videollama}, VideoChat~\cite{li2023videochat}, PandaGPT~\cite{su2023pandagpt}, Video-ChatGPT~\cite{maaz2023video}, BuboGPT~\cite{zhao2023bubogpt}, Valley~\cite{luo2023valley}, VideoChat2~\cite{li2024mvbench}, and Video-LLaVA~\cite{lin2023video}. This type of model processes all modalities, including video, by using a combination of encoders and projection modules to obtain modality features, which are then collectively input into the LLM for further processing. The advantage of this method is its lower training parameter requirements, which can quickly enhance the model's ability to handle multimodal data. Models using the second type of interaction include Otter~\cite{li2023mimic}, MultiModal-GPT~\cite{gong2023multimodal}, MACAW-LLM~\cite{lyu2023macaw}, LLaMA-VID~\cite{li2023llama}, and X-InstructBLIP~\cite{panagopoulou2023x}. These models perform initial alignment and interaction of modality features using structures like cross-attention layers or QFormers~\cite{li2023blip} before the LLM, thereby reducing the subsequent feature processing burden on the LLM. However, these models have higher training costs compared to the first type. Furthermore, models that need to handle complex video understanding tasks, visual grounding modules are typically introduced to deeply understand the positional relationships between objects in the video frames, as seen in PG-Video-LLaVA and BuboGPT. To better process image sequences in videos, models like PG-Video-LLaVA, Video-LLaMA-2, Video-ChatGPT, and Valley extract and integrate both temporal and spatial features of the image sequences, resulting in improved video feature representation.

To thoroughly introduce the aforementioned two types of interaction methods, we selected the Video-LLaMA model from the first type and the X-InstructBLIP model from the second type. We conducted a detailed analysis from four aspects: model architecture, dataset usage, training methods, and evaluation results.

\textbf{Video-LLaMA}


\textbf{Architecture}: Video-LLaMA is designed with two branches: the visual-language branch and the audio-language branch, which respectively convert video frames and audio signals into query representations compatible with LLM text input. In the visual-language branch, video frames are first processed by the BLIP2~\cite{li2023blip} image encoder to obtain image features for each frame. These features are then added to their corresponding position embeddings and fed into the video Q-Former. The video Q-Former extracts features from all frames and outputs a fixed-length comprehensive representation. Finally, a linear layer maps the video features to the same dimension as the text features in the subsequent LLM.
In the audio-language branch, the audio signal is divided into segments of 2 seconds each. Each segment is then converted into a 128-dimensional Mel-spectrogram feature. The encoder from the multimodal model ImageBind~\cite{girdhar2023imagebind} is used to extract features from these audio segments. Similar to the visual-language branch, the audio features are combined with their corresponding position embeddings and fed into the Audio Q-Former. The Audio Q-Former, like the Video Q-Former, maps the features of all audio segments into a fixed-length comprehensive representation. Finally, a linear layer maps this feature to the same dimension as the LLM features.

\textbf{Datasets}: For the vision-language branch, Video-LLaMA is pre-trained on the video-text dataset Webvid-2M~\cite{bain2021frozen} and the image-caption dataset CC595K~\cite{sharma2018conceptual,liu2024visual}. Subsequently, it is fine-tuned using the instruction datasets from MiniGPT-4~\cite{zhu2023minigpt}, LLaVA~\cite{liu2024visual}, and VideoChat~\cite{li2023videochat} to enhance the model's video understanding and instruction comprehension capabilities.
For the audio-language branch, due to the lack of large-scale, high-quality audio-text datasets, this work leveraged the strong cross-modal processing capabilities of ImageBind. It trained the audio-language branch directly using visual-text datasets. Despite not being trained on audio-text data, ImageBind's robust alignment capabilities allowed the Video-LLaMA to effectively align audio to the LLM's textual space during testing.

\textbf{Training \& Evaluation}: The modules involved in the training of Video-LLaMA include the Video Q-Former, Audio Q-Former, the position embeddings, and linear layers in both branches. The image encoder, audio encoder, and the final LLM is frozen during training. This work only provides some test examples without presenting quantitative metrics for the Video-LLaMA.

\textbf{X-InstructBLIP}

\textbf{Architecture}: X-InstructBLIP~\cite{panagopoulou2023x} uses the Q-Former from BLIP2 to process all non-text data. After extracting features for each modality, the model employs a Q-Former with two transformer layers to compute cross-attention between modalities, facilitating interaction between multimodal features. The Q-Former responsible for interaction then outputs a fixed-length comprehensive feature, which is mapped to the feature space of the LLM through a projection layer. Within the LLM, this feature further interacts with the instruction text features to produce the final response.

\textbf{Datasets}: The model's video encoder is initialized with parameters from the image encoder trained on the COCO caption dataset~\cite{changpinyo2021conceptual}. Additionally, it was trained on datasets such as WebVid-2M~\cite{bain2021frozen} and MSRVTT QA~\cite{xu2017video} to enhance its video captioning and video question-answering capabilities. To better assess the model's cross-modal reasoning abilities, this work introduces DisCRn, the first dataset designed for evaluating instruction-based cross-modal discriminative reasoning.

\textbf{Training \& Evaluation}: During training, X-InstructBLIP freezes the parameters of the Q-Former used for extracting modality features and the top LLM. Only the Q-Former responsible for modal interaction and the projection layer are trained. Furthermore, X-InstructBLIP outperforms other models in test tasks across 3D, audio, image, and silent video modalities.

\subsubsection{Video Generation Task}
\ 
\newline
\textbf{Task Description}: 

Video generation refers to the training of artificial intelligence to autonomously produce high-fidelity video content that aligns with given descriptions using single-modal or multimodal data such as text, images, or videos. Based on the method of generation, current AI video generation can be categorized into text-conditioned video generation, image-conditioned video generation, and video-to-video synthesis. Specifically, the text-conditioned video generation task generates videos based on textual instructions. In the image-conditioned video generation task, the model generates corresponding videos from one or more input reference images. The video-to-video synthesis task generates a new video from an input video. With the development of MLLMs, more models now support any-to-any tasks~\cite{wu2023next,tang2024any,zhan2024anygpt}, allowing for conversions and generation across various modalities, including using text, images, or videos simultaneously to guide the generation of new videos.

\noindent{\textbf{Model Introduction}:}

The technological evolution of video generation can be broadly divided into three stages, including deep learning based models, autoregressive models, and diffusion models. Early on, the introduction of Generative Adversarial Networks~\cite{goodfellow2020generative} (GANs) and Variational Autoencoders~\cite{kingma2013auto} (VAEs) marked significant turning points in video generation technology. However, these deep learning based methods are typically singular, static, and of lower resolution. Later, the transformer-based autoregressive models, such as Video Transformer~\cite{neimark2021video}, improved video generation performance and efficiency by capturing long-range dependencies in video sequences through self-attention mechanisms. After that, diffusion models~\cite{ho2022video} have become the mainstream technological path in AI video generation. Compared to previous generation models like GANs, the training process of diffusion models is typically more stable, resulting in higher-quality images or videos, especially when well-trained, the generated results are often more realistic. Additionally, diffusion models do not rely on specific network structures, exhibiting good compatibility.

To achieve LLMs with multimodal inputs and outputs, some researchers are exploring using LLMs as decision-makers and leveraging existing pre-trained multimodal encoders and decoders as tools for processing multimodal inputs and outputs, such as Visual-ChatGPT~\cite{wu2023visual}, HuggingGPT~\cite{shen2024hugginggpt}, and AudioGPT~\cite{huang2024audiogpt}. However, the output of these models heavily depends on the quality of the chosen decoder and has limited controllability. To achieve better generation results, subsequent methods incorporate decoders into the model training framework. For instance, NeXT-GPT~\cite{wu2023next} includes projection modules between the encoder and LLM and between the decoder and LLM. The projection module between the encoder and LLM, similar to those in video understanding models, ensures accurate alignment of each modality’s features to the LLM's feature space. The projection module between the decoder and LLM maps the LLM's output features into representations understandable by each decoder. In the Vitron~\cite{fei2024vitron} model, researchers directly train the LLM to produce output features better suited for the decoder. The EMO~\cite{tian2024emo} model uses a VAE to generate dynamic speaking videos based on a reference facial image and voice audio. There is another category of models that do not use LLMs, such as the Large World Model~\cite{liu2024world} based on autoregressive models, which can generate subsequent feature representations based on prior input modality information. During training, these models first train multimodal alignment using image caption datasets, followed by instruction tuning with various data to generate more compliant videos. Given the representative architecture and training methods of the NeXT-GPT model, we provide a detailed introduction of NeXT-GPT in terms of model architecture, dataset usage, training methods, and evaluation results.

\textbf{NeXT-GPT}



\textbf{Architecture}: NeXT-GPT uses ImageBind to extract features from various modalities. Each modality's features are then mapped to the LLM space via their respective projection layers. The projected features, along with the instruction text, are input into an LLM, specifically Vicuna (7B-v0)~\cite{zheng2024judging}. After understanding and reasoning by the LLM, the output features for each modality are obtained. To better adapt to the subsequent generation process, NeXT-GPT utilizes a Transformer-based projection layer before each multimodal decoder, converting the signal token representations from the LLM into features more suitable for the decoder. Notably, the projection layers in both the encoding and decoding stages are Transformer-based. The model also incorporates learnable concept tokens, which hierarchically aggregate grid-level features into semantic concept tokens through a grouping mechanism. These conceptual representations are then fed into the LLM.

\textbf{Datasets}: To train multimodal alignment capabilities, the model primarily used Webvid-2M~\cite{bain2021frozen}, CC3M~\cite{sharma2018conceptual}, and AudioCaps~\cite{kim2019audiocaps} as training data for video-caption, image-caption, and audio-caption tasks, respectively. NeXT-GPT also constructed the MosIT instruction datasets. This dataset not only supports complex cross-modal understanding and reasoning tasks but also facilitates higher-quality multimodal data generation. Specifically, it is designed to include multi-turn dialogues, each consisting of 3-7 interactions, involving various cross-modal interactions.

\textbf{Training \& Evaluation}: During training, NeXT-GPT keeps the encoders and decoders for each modality and the intermediate LLM frozen. Only the Transformer-based projection layers in the encoding and decoding stages are trainable, achieving lightweight multimodal alignment learning. Additionally, to enhance the capabilities and controllability of the LLM, NeXT-GPT uses the self-constructed MosIT dataset for modality-switching instruction tuning. During instruction tuning, besides updating the projection layers, this model also applies LoRA~\cite{hu2021lora} to further update the LLM's parameters. For evaluating video understanding tasks, the method tests the model's video captioning and video question answering capabilities on MSRVTT~\cite{xu2016msr}, MSVD-QA~\cite{xu2017video}, MSRVTT-QA~\cite{xu2017video}, and NExTQA~\cite{xiao2021next} datasets. Additionally, this work tests the model's video generation ability on the MSRVTT dataset. NEXT-GPT demonstrates excellent performance in these evaluations.

\subsection{Audio Tasks}
MLLM has made rapid progress in the field of artificial intelligence research, however, there are still some limitations in audio modality-related tasks. As one of the essential information modalities in the human world, the understanding and generation of audio is a necessary path for the development of higher artificial intelligence, therefore, the development and application of MLLM in the field of audio still has a broad prospect and necessity.

Audio processing technology is constantly updated and iterated with the development of information technology~\cite{willmore2023adaptation}. Before the emergence of the LLM, audio processing technology experienced traditional methods such as template matching and the Hidden Markov Model (HMM), as well as DNN, CNN, LSTM, and other methods based on deep learning~\cite{bohnenstiehl2023automated,deshmukh2023pengi,pan2024research,shafieian2023hidden}. These methods laid the foundation for the development of speech technology. With the development of self-supervised learning (SSL), many speech processing problems have been effectively solved, e.g., HuBERT uses masked prediction training~\cite{hsu2021hubert}, SoundStream proposes a high-level hierarchical architecture with semantic information~\cite{zeghidour2021soundstream}. DiscreTalk constructs speech synthesis based on the Discrete Sequence Generation model, VQ-VAE, and the Autoregressive Transformer-NMT model~\cite{hayashi2020discretalk}, textless-lib builds speech models directly by training an autoregressive generative model for low bit rate audio tokens~\cite{kharitonov2022textless}, and so on.

After the emergence of ChatGPT, many researches began to shift their focus to the integration of speech models with LLM, hoping to realize direct speech interaction between existing speech models and LLM. The advantages of multimodal speech models in speech tasks are significant compared to traditional methods of speech task implementation. Compared with single-modal models, multimodal speech models have cross-modal alignment capability, greater versatility and flexibility, and can realize the interaction and transformation of audio modality with multiple modalities such as text, image, video, etc~\cite{zhan2024anygpt,shen2024hugginggpt}. The multimodal speech model has better semantic understanding and generation ability, and the introduction of LLM makes the model have more powerful multimodal perception and understanding ability, and the multimodal speech model improves the understanding and execution of the task commands such as text and speech significantly compared with the traditional speech processing methods.

In this section, we will introduce the two broad categories of audio comprehension and audio generation, followed by the specific tasks and related MLLM in audio comprehension and audio generation, respectively.

\subsubsection{Audio Understanding}

\ 
\newline
\textbf{Task Description}: 

In audio understanding, MLLM has demonstrated powerful capabilities in tasks such as automatic speech recognition (ASR)~\cite{malik2021automatic}, and speech-to-text translation (S2TT)~\cite{bahar2019comparative}. These diverse applications underscore the versatile and robust capabilities of MLLMs in the domain of audio understanding.

\ 
\newline
\textbf{Model Introduction}: 

Many advances have been made in realizing audio understanding tasks by constructing large-scale datasets for instruction fine-tuning.
For example, in the generalized task of audio comprehension class, Qwen-Audio developed Qwen-Audio-chat by instruction fine-tuning to support multi-task training, solved the problem of transforming text labels in different datasets, and extended the training to dozens of datasets covering more than 30 tasks, 8 languages, and various types of audio in order to improve generalized audio comprehension~\cite{chu2023qwen}. Similarly, MACAW-LLM, to address the limitations of current multimodal datasets that primarily emphasize specific task types, created the MACAW-LLM instruction dataset that covers a wide range of instructional tasks and combines a variety of data modalities to make it more varied and suitable for multimodal instruction-tuned LLMs~\cite{lyu2023macaw}. The superior generative capabilities of the current LLMs were utilized to manage this dataset to ensure that the target text is correctly aligned with human instructions. Gemini trains several modalities, including text, image, audio, and video, from scratch, using a Decoder-only model structure, optimized for structural and optimization goals to improve the stability of large-scale training and inference, with speech recognition and speech translation capabilities in audio tasks~\cite{team2023gemini}.

There has also been a lot of work on multimodal-aligned instruction fine-tuning methods that have made significant progress. For example, in speech recognition and translation tasks, AudioPaLM fuses the text-based and speech-based language models PaLM-2 and AudioLM into a unified multimodal architecture that achieves feature fusion by taking a pre-trained text-only model and extending its embedding matrix to model a new set of audio tokens~\cite{rubenstein2023audiopalm}. For automatic speech recognition and audio captioning tasks, SALMONN uses a dual-encoder architecture, linking cross-modal modules via Q-former and low-rank adaptation (LoRA) methods, and aligning the input space with the output space through pre-training and instruction tuning~\cite{tang2023salmonn}. For audio-visual understanding tasks, Video-LLaMA bootstraps cross-modal training from frozen pre-trained visual and audio encoders as well as frozen LLMs, utilizes pre-trained audio encoders and audio Q-former to learn reasonable auditory query embeddings, and fine-tunes the alignment of the different modalities into a common space through a series of steps to achieve the alignment effect of each modality~\cite{cheng2024videollama}. Similarly, BuboGPT~\cite{zhao2023bubogpt}.

Next, we provide an in-depth understanding of the contribution of MLLM to audio comprehension tasks by analyzing Qwen-Audio and SALMONN. Qwen-Audio achieves multitask training covering a wide range of audio comprehension tasks and improves generalized audio comprehension by constructing a large-scale dataset and instruction fine-tuning, while SALMONN improves generalized audio comprehension by using a dual-encoder architecture and cross connection of modal modules, which realizes the alignment of input space and output space, provides an effective solution for automatic speech recognition and audio subtitle generation tasks.

\textbf{Qwen-Audio}: 

\textbf{Model Architecture}: Qwen-Audio contains an audio encoder and a . The audio encoder uses the Whisper-large-v2 model of the 32-layer Transformer as the initialization model, and Qwen-7B serves as the base component of the . The training goal of the model is to maximize the probability of the next text token given pairs of audio sequences and text sequences.

\textbf{Datasets}: To address the challenge of co-training multiple tasks and multiple datasets, Qwen-Audio proposes a multi-task training framework. The framework conditions the output of the decoder as a series of hierarchical labels on data from more than 30 tasks, eight languages, and multiple audio types, mitigating differences in the multitasking data due to task goals, languages, annotation granularity, and text structure by sharing labels. In addition, to better understand speech signals in tasks such as speech recognition and audio Q\&A, fine-grained word-level timestamps are introduced in the training to improve the model's ability to align audio signals and timestamps.

\textbf{Training \& Evaluation}: By unifying the training format for multi-task datasets, the Qwen-Audio model improves performance by maximizing knowledge sharing between similar tasks. During training, only the audio encoder is optimized, and the  is optimized by instruction fine-tuning after training is complete. With this two-stage training, Qwen-Audio demonstrates strong generalized comprehension across multiple tasks.

\textbf{SALMONN}:

\textbf{Model Architecture}: SALMONN consists of two components designed to achieve audio-text alignment with high temporal resolution. First, the auditory signal processing uses a dual-encoder architecture: a Whisper-Large-v2 based speech encoder is responsible for encoding the speech signal, and a fine-tuned BEATs-based audio encoder is responsible for encoding the non-speech signal. The outputs of the two auditory encoders are connected via a window-level Query Transformer in order to transform the encoder outputs into an augmented audio signal for input to the Vicuna LLM. Next, the enhanced LLM input and output are instruction tuning by a Low-Rank Adaptive Approach (LoRA) to achieve audio-text cross-modal alignment, thus endowing the model with zero-shot capability.

\textbf{Datasets}: Speech recognition datasets LibriSpeech and GigaSpeech M-sets, and audio captioning datasets WavCaps and AudioCaps are used as pre-training datasets, and datasets under multiple tasks are used for instruction fine-tuning.

\textbf{Training \& Evaluation}: In the pre-training stage, Q-Former and LoRA are trained using the above auditory datasets to learn the alignment between audio signals and text. In the second phase, instruction tuning is performed through a series of tasks aimed at addressing the overfitting problem of the model. The instruction-tuned model was evaluated on the following three levels of tasks: tuned tasks (tasks that the model had seen during instruction tuning), speech-based NLP tasks ( such as speech-to-text conversion, speech emotion recognition, etc.), and speech-based and non-speech-based comprehension tasks (such as speech-audio collaborative reasoning, etc.). The evaluation results show that the model performs competitively in all of the above tasks, validating its effectiveness in high temporal resolution audio-text alignment and cross-modal alignment.

\subsubsection{Audio Generation}

\ 
\newline
\textbf{Task Description}: 

Audio generation involves training models to accept multiple input modalities, such as text, speech, images, and more, to understand these multimodal contents and create, synthesize, or improve compliant audio outputs. This task is crucial in the construction of multimodal datasets and instruction fine-tuning. By investigating the application of multimodal large language models to audio generation, we can better understand their working mechanisms and performance.   

\ 
\newline
\textbf{Model Introduction}: 

SpeechGPT~\cite{zhang2023speechgpt} and AnyGPT~\cite{zhan2024anygpt} are two representative models in the construction of multimodal datasets. SpeechGPT, through its discrete representation and training on cross-modal instruction datasets, excels not only in text-to-speech transformation (TTS) tasks but also in speech generation and dialog capabilities. Similarly, AnyGPT is trained with discrete representations and cross-modal instruction datasets. As an ``any to any" multimodal model, AnyGPT can transform and generate between different modalities, achieving high-quality transformation from various input modalities to audio output. By utilizing discrete representations and cross-modal dataset training, SpeechGPT and AnyGPT demonstrate robust generative capabilities in TTS tasks, laying a solid foundation for audio generation. 

In terms of multimodal alignment and modal bridging, AudioGPT~\cite{huang2024audiogpt} and HuggingGPT~\cite{shen2024hugginggpt} make significant contributions. AudioGPT is based on a cascade architecture and is connected to LLM, enabling it to call the corresponding base model to complete speech output in TTS tasks. AudioGPT performs exceptionally well in multi-round conversations, handling complex scenarios and generating coherent and natural speech. HuggingGPT first utilizes LLMs to decompose human commands and then calls Huggingface's models to accomplish specific tasks. It covers a range of models for handling speech tasks, demonstrating flexibility to meet different audio generation needs and excelling in handling multimodal instructions. Through multimodal alignment and modal bridging, AudioGPT and HuggingGPT show excellent performance in TTS tasks and multi-round conversations, providing more flexible and efficient solutions for audio generation. 

Through the construction of multimodal datasets and multimodal alignment and bridging, MLLM enables precise control over audio generation and produces high-quality outputs. SpeechGPT, AnyGPT, AudioGPT, and HuggingGPT demonstrate strong generative capabilities and command adherence in their respective application scenarios, highlighting the great potential and broad prospects of multimodal large language models in audio generation. The successful application of these models provides a crucial theoretical foundation and practical experience for the future development of audio generation technology. To fully understand the contributions of multimodal large language models in audio generation tasks, we will introduce the models in detail, focusing on SpeechGPT and AudioGPT

\textbf{SpeechGPT}: 

\textbf{Model Architecture}:The model consists of three main components: discrete unit extractor, LLM, and unit vocoder. discrete unit extractor uses Hidden-unit BERT (HuBERT) to convert continuous speech signals into a sequence of discrete units, LLM uses the Meta AI LLaMA model to sense multimodal inputs and generate multimodal outputs, and unit vocoder is multi-speaker unit Hi-Fi-GAN to decode discrete representations into speech signals.

\textbf{Datasets}: LibriLight as modality-adaption pre-training stage dataset, gigasspeech, common voice, librisspeech, and moss-002-sft-data as cross-modal instruction fine-tuning stage dataset and moss-002-sft-data as chain-of-modality instruction fine-tuning stage.

\textbf{Training \& Evaluation}: The training process is divided into three stages: first, the Modality-Adaption Pre-training Stage is performed, in which discrete unit signal-text pairs are constructed based on the LibriLight dataset. Then the Cross-Modal Instruction Fine-tuning Stage is performed, in which the cross-modal instruction dataset, i.e., the Multi-Modal Instruction-Discrete Unit-Text dataset, is constructed on the basis of the existing discrete unit-signal-text pairs, which are generated from the task descriptions generated by GPT-4. Finally, the Chain-of-Modality Instruction Fine-tuning Stage is performed to train the Text-Discrete Unit Generator to transform the text instructions into speech instruction data to obtain Speech Instruction-Speech Response, Speech Instruction- Text Response, Text Instruction-Speech Response, Text Instruction-Speech Response, and Text Instruction-Text Response chained instructions, which are used to realize a MLLM with intrinsic cross-modal conversational capabilities.

\textbf{AudioGPT}:

\textbf{Model Architecture}: first is the modality transformation, which is used to transform the various modalities of the input into a query with consistent modalities; second is the LLM module, in which ChatGPT is implanted, used for task analysis by the dialog engine together with the cue manager; the third is responsible for the assignment of the task to the appropriate audio base model; and finally, the generation of the response after the execution of the task.

\textbf{Datasets}: A variety of datasets for use under the audio model task are provided to validate the ability of audioGPT to realize generic audio tasks through the audio base model.

\textbf{Training \& Evaluation}:  The process is divided into four main phases: modality transformation, task analysis, modality transformation, and response generation. The modal shift phase uses an input/output interface to shift between speech and text, bridging the gap between the language model LLM and ChatGPT. The task analysis phase uses the dialog engine and cue manager to help ChatGPT understand the user's intention to process audio information. In the model assignment phase, ChatGPT receives structured inputs such as rhymes, timbres, and speech controls, and assigns appropriate audio base models for comprehension and generation. In the Response Generation phase, the audio base model performs the task of generating and returning the final response for the user.

\section{Comparison of MLLMs}

\subsection{Image Tasks}

MLLMs in the image domain have made remarkable progress in recent years, demonstrating superior performance on complex visual tasks. Table \ref{tab:mlm_models_specifications_img} provides an overview and comparison of several representative image MLLMs, focusing on their innovations and advantages in architecture design, multimodal fusion, dataset selection, and downstream task adaptation.

In terms of architecture design, image MLLMs exhibit a diverse trend. On one hand, many models adopt the classic ``dual-tower" structure, which aligns pre-trained language models and vision models in parallel, achieving cross-modal information fusion through alignment modules. For example, the LLaVA~\cite{chen2024llava} series models employ the Vicuna language model and CLIP vision model, performing alignment via simple linear layers, and have achieved exceptional performance in tasks such as image classification and image-text generation. On the other hand, some models explore more intimate fusion approaches. BLIP-2~\cite{li2023blip}, for instance, utilizes fixed image/text encoders and achieves deep interaction through the QFormer alignment module, exhibiting strong comprehension capabilities in visual question answering and image description tasks. Furthermore, there are models that attempt to introduce additional modalities. VPGTrans~\cite{zhang2024vpgtrans}, for example, combines visual, linguistic, and positional information, achieving more comprehensive image understanding through cross-modal attention mechanisms.

Multimodal fusion techniques are at the core of image MLLMs, and their design directly impacts the models' performance and efficiency. In addition to common linear layers and attention mechanisms, some models introduce novel fusion approaches. For instance, MultiModal-GPT~\cite{gong2023multimodal} employs a Transformer-based fusion module, achieving deep interaction between text and images through self-attention and cross-attention. OpenFlamingo~\cite{awadalla2023openflamingo} adopts a progressive fusion strategy, gradually integrating multimodal information in different Transformer blocks, enhancing the model's representational power. Moreover, some models explore fusion methods based on graph neural networks. ALLaVA~\cite{chen2024allava}, for example, captures structured information between modalities by constructing relation graphs of text and images. These innovative fusion techniques provide new perspectives for further improving the performance of image MLLMs.

Data is the foundation of model training, and the development of image MLLMs relies on high-quality, large-scale multimodal datasets. Besides commonly used image-text datasets (e.g., LAION, Conceptual Captions) and image annotation datasets (e.g., COCO, Visual Genome), some models introduce more diverse data sources. For instance, LLaVA-1.5~\cite{liu2024improved} leverages open-domain visual question answering and Optical Character Recognition (OCR) data, improving the model's comprehension ability in open-domain scenarios. Otter utilizes the MIMIC-IT~\cite{li2023mimic} multimodal dataset from the medical domain, showcasing the application potential of image MLLMs in vertical domains. In the future, building larger-scale, higher-quality image datasets covering a wider range of domains will be key to driving the continuous development of this field.

The adaptation and transfer capabilities of image MLLMs on downstream tasks have received significant attention. Some models employ universal instruction fine-tuning methods. For example, MiniGPT-4~\cite{zhu2023minigpt} and LLaVA ~\cite{chen2024llava} achieve rapid adaptation to tasks such as image classification, object detection, and image-text generation by fine-tuning on task-specific data. Other models explore more efficient transfer learning methods. LLaMA-Adapter V2~\cite{gao2023llama}, for instance, achieves fast adaptation to downstream tasks through learnable adapter modules while keeping the pre-trained model fixed. Yi-VL~\cite{young2024yi} introduces knowledge extraction and enhancement techniques, improving the model's performance on open-domain tasks by acquiring task-relevant information from external knowledge bases. These adaptation and transfer methods tailored for downstream tasks provide valuable references for the flexible deployment of image MLLMs in practical applications.

Through deep fusion of language and vision, these models have demonstrated performance surpassing single-modal models in tasks such as image understanding, question answering, and generation. In the future, research on image MLLMs will focus on optimizing model architectures, innovating multimodal fusion, constructing large-scale datasets, flexibly adapting to downstream tasks, and addressing challenges in interpretability, fairness, and privacy protection. As research continues to deepen, image MLLMs are expected to find widespread applications in areas such as smart cities, autonomous driving, and medical imaging, providing more intelligent and efficient assistance for human cognition and decision-making. At the same time, the development of this field will also drive significant breakthroughs in multimodal perception, reasoning, and interaction in artificial intelligence, bringing more intelligent, natural, and human-centric human-machine collaboration experiences.

\begin{sidewaystable}[thp]
    \centering
    \caption{Summary of MLLMs on Image Tasks}
    \resizebox{\textwidth}{!}{%
    \begin{tabular}{llllllll}
        \toprule
        \textbf{MLLM} & \textbf{Base Model} & \textbf{Architecture} & \textbf{Dataset} & \textbf{Date} & \textbf{Hardware} & \textbf{Parameters} & \textbf{Alignment Module} \\
        \midrule
        BLIP-2 & BLIP & frozen image/text encoder & COCO, Visual Genome, CC3M, CC12M, SBU, LAION400M & 2023.6 & 16*A100 & 1.2B & QFormer \\
        X-InstructBLIP & Vicuna & EVA-CLIP-ViT-G/14, BEATsiter3+, ULIP-2 & DisCRn, MSRVTT, WebVid2M, MSRVTT QA & 2023.11 & 8*A100 & 7B/13B & QFormer \\
        LLaVA & Vicuna & CLIP ViT-L/14 & CC-3M, lava\_instruct\_150k.json, llava\_instruct\_80k.json, conversation\_58k.json, detail\_23k.json, complex\_reasoning\_77k.json & 2023.12 & 8*A100 & 7B/13B & Linear Layer \\
        LLaVA-1.5 & LLaVA & CLIP-ViT-L-336px & openknowledge VQA and OCR OCRVQA TextCaps & 2023.11 & 8*A100 & 7B & Linear Layer \\
        LLaVA-1.6 & LLaVA & - & - & 2023.12 & 8*A100 & 34B & Linear Layer \\
        MiniGPT-4 & Vicuna/LLaMA & BLIP-2, ViT-G/14 & LAION, Conceptual Captions & 2023.4 & 8*A100 & 7B/13B & Linear Layer \\
        MiniGPT-v2 & LLaMA & BLIP-2, ViT-G/14 & LAION, CC3M, SBU, GRIT-20M, COCO caption, RefCOCO, RefCOCO+ & 2023.11 & 8*A100 & 7B/13B & Linear Layer \\
        VPGTrans & Vicuna/LLaMA & BLIP-2, CLIP ViT-L/14 & COCO caption, NoCaps, VQAv2, GQA, and OK-VQA & 2023.4 & 8*A100 & 7B & Linear Layer \\
        MultiModal-GPT & OpenFlamingo & CLIP ViT-L/14/LLaMA & Dolly 15k, Alpaca GPT4, COCO Caption, A-OKVQA & 2023.6 & 8*A100 & 9B & transformer \\
        Otter & LLaMA & CLIP ViT-L/14 & MIMIC-IT & 2023.5 & 4*3090 & 9B & Linear Layer \\
        OpenFlamingo & MPT & CLIP ViT-L/14 & ALIGN, LAION-2, M3W, Multimodal C4 & 2023.8 & 64*A100 & 3B/4B/9B & Linear Layer \\
        LLaMA-Adapter V2 & LLaMA & CLIP ViT-L/14 & GPT4-LLM, COCO & 2023.4 & 8*A100 & 1.2M & Linear Layer \\
        mPLUG-Owl-2 & LLaMA & CLIP ViT-L/14 & LAION, COYO, Conceptual Captions, MSCOCO, Alpaca, Vicuna, Baize & 2024.5 & - & 8.2B/9.8B & Linear Layer \\
        ALLaVA & Phi22/StableLM-2/Phi-3-mini & CLIP ViT-L/14 & LAION, Vision-FLAN & 2024.6 & 8*A100 & 3B & Linear Layer \\
        Yi-VL & LLaMA & CLIP ViT-L/14 & - & 2024.5 & 128*A100 & 6B/9B/34B & Linear Layer \\
        Qwen-VL-Chat & Qwen & CLIP ViT-L/14 & LAION, LAION-COCO, DataComp, Coyo, CC12, CC3M, SBU, COCO Caption & 2023.8 & 1*A100 & 7B & Linear Layer \\
        VILA & LLaMA & CLIP ViT-L/14 & MMC4, COYO & 2024.5 & 16*A100 & 3B/13B/40B & Linear Layer \\
        \bottomrule
    \end{tabular}%
    }
    \label{tab:mlm_models_specifications_img}
\end{sidewaystable}

\begin{sidewaystable}[thp]
    \centering
    \caption{Summary of MLLMs on Video Understanding.}
    \resizebox{\textwidth}{!}{
    \begin{tabular}{llllllll}
        \toprule
        \textbf{MLLM} & \textbf{Language Model} & \textbf{Architecture} & \textbf{Dataset} & \textbf{Time} & \textbf{Parameters} & \textbf{Alignment Module} & \textbf{Hardware} \\
        \midrule
        VideoChat & T5 & InternVideo, Whisper & COCO Caption, Visual Genome, CC3M, CC12M & 2023.5 & 10M & Linear Layer & - \\
        MM-GPT & LLaMA & CLIP ViT & Dolly 15k, Alpaca GPT4, LLaVA, Mini-GPT4, A-OKVQA, COCO Caption, OCR VQA & 2023.5 & 9B & Linear Layer & 8*A100 \\
        PandaGPT & Vicuna & ImageBind & LLaVa, Mini-GPT4 & 2023.5 & 13B & Linear Layer & 8*A100 \\
        Otter & LLaMA & CLIP ViT-L/14 & MIMIC-IT & 2023.5 & 9B & Linear Layer & 4*3090 \\
        MACAW-LLM & LLAMA & CLIP-VIT-B/16, WHISPER & Alpaca instruction dataset, COCO-caption, Charades, AVSD & 2023.6 & 7B & Linear Layer & 8*A100 \\
        Video-ChatGPT & Vicuna & CLIP ViT-L/14 & ActivityNet-200 & 2023.6 & 7B & Linear Layer & 8*A100 \\
        Video-LLaMA & ImageBind & BLIP2 & Webvid-2M, CC595K & 2023.6 & 7B/13B & QFormer & 8*A100 \\
        BuboGPT & Vicuna & ImageBind, BLIP-2 & CC3M, CC12M, SBU, LAION, WaveCaps, LLaVA data & 2023.7 & 7B & QFormer & - \\
        Valley & Vicuna & CLIP ViT-L/14 & CC3M, WebVid2M & 2023.10 & - & Linear Layer & 8*A100 \\
        VideoChat2 & Vicuna & UMT-L & MVBench, CC3M, CC12M & 2023.11 & 7B & QFormer & 32*A100 \\
        LLaMA-VID & Vicuna & EVA-G & CC3M, WebVid 2.5M, ShareGPT, LongLoRA, MovieNet & 2023.11 & 7B & Linear Layer & 8*A100 \\
        X-InstructBLIP & Vicuna & EVA-CLIP-ViT-G/14, BEATsiter3+, ULIP-2 & DisCRn, MSRVTT, WebVid2M, MSRVTT QA & 2023.11 & 7B/13B & QFormer & 8*A100 \\
        Video-LLaVA & Vicuna & LanguageBind & LAION-CCSBU, CC3M, WebVi & 2023.11 & 7B & Linear Layer & 8*A100 \\
        Gemini & - & - & - & 2023.12 & - & Transformers & TPUv4 \& TPUv5e \\
        PG-Video-LLaVA & Vicuna & Vicuna-1.5, CLIP-ViT-L-14 & VideoInstruct100K & 2023.12 & 7B/13B & Linear Layer & 4*A100 \\
        VILA & Vicuna & Vicuna-1.5-7B/13B, CLIP-ViT-L-14 & MMC4, COYO, TextCaps, ActivityNet-QA & 2023.12 & 7B/13B & Linear Layer & 16*A100 \\
        Mantis & LLaMA 3 & Idefics2 & VIST, NExT-QA, STAR & 2024.5 & 8B/9B & Linear Layer & 16*A100 \\
        LLaVA-NeXT-Video (7B) & Vicuna & CLIP-ViT-L-14 & LAION-GPT-V, ShareGPT-4V, NExT-QA, LLaVA-Hound & 2024.5 & 7B/34B & Linear Layer & 32*A100 \\
        Video-LLaMA-2 & Mixtral-Instruct & CLIP ViT-L/14, BEATs & Panda-70M, VIDAL-10M, WebVid-10M, InternVid-10M & 2024.6 & 7B/56B & Linear Layer & 8*A100 \\
        \bottomrule
    \end{tabular}
    }
    
    \label{tab:mlm_models}
\end{sidewaystable}

\subsection{Video Understanding}

Table~\ref{tab:mlm_models} summarizes some representative MLLMs and their characteristics. These models are primarily based on pre-trained language models such as T5, LLaMA~\cite{touvron2023llama}
, and Vicuna~\cite{zheng2024judging}, and are integrated with different visual models like CLIP, ImageBind, and EVA. For instance, MM-GPT employs the LLaMA language model and CLIP ViT visual model, and is pre-trained on large-scale multimodal datasets such as Dolly 15k and Alpaca GPT4, resulting in a model with 9 billion parameters. Similarly, PandaGPT~\cite{su2023pandagpt} utilizes the Vicuna language model and ImageBind visual model, trained on datasets like LLaVa and Mini-GPT4~\cite{zhu2023minigpt,chen2023minigptv2}, with a parameter count of 13 billion. Although these large-scale MLLMs have achieved exceptional performance on downstream tasks such as video question answering and video description, they also face challenges like high training costs and slow inference speeds. Striking a balance between model performance and computational efficiency is a critical issue that needs to be addressed.

Multimodal fusion is a key component of MLLMs, responsible for mapping visual and language features into a common semantic space. The current mainstream fusion methods include simple linear layers and attention mechanisms (such as QFormer). Compared to linear layers, attention mechanisms can capture more complex interactions between visual and language features, but also introduce higher computational overhead. Some of the latest models, such as LLaVA-NeXT-Video~\cite{zhang2024llavanext-video} and Video-LLaMA-2~\cite{cheng2024videollama}, attempt to introduce more advanced multimodal fusion modules, such as Self-Attention and Cross-Attention, further enhancing model performance. Innovations in multimodal fusion techniques will be a crucial driving force for the development of MLLMs.

Despite the impressive results achieved by MLLMs in video understanding benchmarks, applying them to real-world scenarios still faces numerous challenges. Firstly, the annotation quality of large-scale video datasets varies greatly, which can introduce noise and affect model performance. Obtaining high-quality video annotation data is an urgent problem that needs to be addressed. Secondly, current MLLMs have parameters in the order of billions, resulting in slow inference speeds, making it difficult to meet the requirements of real-time applications. Techniques like model quantization and pruning may help alleviate this issue. Moreover, existing MLLMs are sensitive to interference factors in videos, such as changes in lighting, occlusion, and blur. Improving the robustness of models is crucial for practical applications. Lastly, MLLMs are typical ``black-box models" that lack interpretability, which may raise safety concerns in high-risk application domains (e.g., autonomous driving). Developing interpretable MLLMs is an important future research direction.

Looking ahead, there are still many research directions worth exploring for MLLMs in the field of video understanding. Firstly, more efficient visual-language representation learning methods, such as contrastive learning and reinforcement learning, have the potential to further improve model performance. Secondly, more advanced multimodal fusion mechanisms, like Temporal Attention that considers temporal information, can better model the dynamic features of videos. Furthermore, more lightweight model designs that significantly reduce computational overhead while maintaining performance are more suitable for real-time application scenarios. Additionally, MLLMs based on high-level cognitive abilities, such as causal reasoning and logical reasoning, are closer to human-like video understanding. Finally, incorporating external knowledge (e.g., common sense, domain knowledge) to enhance MLLMs, enabling them to possess stronger analysis and reasoning capabilities, is also a promising research direction.

\subsection{Video Generation}
Table \ref{tab:additional_mlm_models} reviews and compares several representative video generation MLLMs, focusing on their architectural design, multimodal fusion, practical applications, and future research directions.

Regarding model architecture, various MLLMs employ diverse designs. For instance, HuggingGPT~\cite{shen2024hugginggpt}  uses a LLM as the core controller, EMO incorporates specialized facial feature extraction and audio processing modules, while LWM~\cite{liu2024world} and NeXT-GPT~\cite{wu2023next} integrate advanced generative adversarial networks (such as VQGAN~\cite{yu2021vector}) and vision-language models (like CLIP and ImageBind~\cite{girdhar2023imagebind}). These architectural choices significantly impact the models' generative capabilities and efficiency. For example, LWM boasts a parameter count of up to 1M and robust scalability, whereas NeXT-GPT achieves superior video quality with smaller parameter sizes (7B/13B).

Multimodal fusion technology is a crucial component of MLLMs, responsible for mapping information from different modalities—such as text, images, and audio—into a shared semantic space. Common fusion methods include simple linear layers (as seen in Vitron~\cite{fei2024vitron}) and attention mechanisms (as utilized in LWM's Transformer structure). Compared to linear layers, attention mechanisms can capture more complex inter-modal interactions, albeit at the cost of higher computational demands. 

Despite the achievements of these MLLMs in video generation tasks, several challenges hinder their application in real-world scenarios. The training and inference processes are highly resource-intensive, often requiring hundreds of high-end GPUs~\cite{laccetti2013high}, which limits the accessibility and practicality of the models. Additionally, there is room for improvement in the quality and diversity of the generated videos, particularly concerning visual coherence, temporal consistency, and detail depiction. 

Looking ahead, the research directions for video generation MLLMs should focus on developing more efficient model architectures and training strategies to reduce computational costs while maintaining performance. Exploring advanced multimodal fusion mechanisms, such as hierarchical fusion and contrastive learning~\cite{giorgi2020declutr}, will enhance semantic alignment across modalities. Incorporating external knowledge, including commonsense and causal relationships, can improve the models' understanding and reasoning capabilities. Enhancing model interpretability and controllability to support user customization and specific constraints is another key area.

\begin{sidewaystable}[thp]
    \centering
    \caption{Summary of MLLMs on Video Generation.}
    \resizebox{\textwidth}{!}{%
    \begin{tabular}{llllllll}
        \toprule
        \textbf{MLLM} & \textbf{Language Model} & \textbf{Architecture} & \textbf{Dataset} & \textbf{Time} & \textbf{Parameters} & \textbf{Alignment Module} & \textbf{Hardware} \\
        \midrule
        HuggingGPT & - & LLM (e.g., ChatGPT) as the core controller & - & 2023.3 & - & - & - \\
        EMO (Emote Portrait Alive) & - & SD 1.5 UNet, ReferenceNet, wav2vec & HDTF, VFHQ & 2024.2 & - & Linear Layer & - \\
        LWM & LLaMA & VQGAN & LAION-2Ben, COYO-700M, WebVid10M, 3M InternVid10M, ShareGPT4V & 2024.2 & 256K/512K/1M & Transformer & 450*A100 \\
        Vitron & Vicuna & CLIP ViT-L/14 & CC3M, Webvid, RefCOCO & 2024.4 & 7B & Linear Layer & 10*A100 \\
        NeXT-GPT & Vicuna & ImageBind & CC3M, COCO-caption, WebVid-2M, AudioCaps, LLaVA-150K, VideoChat & 2023.9 & 7B/13B & Linear Layer & 18*A100 \\
        \bottomrule
    \end{tabular}%
    }
    \label{tab:additional_mlm_models}
\end{sidewaystable}

\begin{sidewaystable}[thp]
    \centering
    \caption{Summary of MLLMs on Audio Tasks.}
    \resizebox{\textwidth}{!}{%
    \begin{tabular}{llllllll}
        \toprule
        \textbf{MLLM} & \textbf{Language Model} & \textbf{Architecture} & \textbf{Dataset} & \textbf{Time} & \textbf{Parameters} & \textbf{Alignment Module} & \textbf{Hardware} \\
        \midrule
        PandaGPT & Vicuna & ImageBind & data from LLaVa, Mini-GPT4 & 2023.5 & 13B & Linear Layer & 8*A100 \\
        Video-LLaMA & ImageBind & BLIP2 & Webvid-2M, CC595K & 2023.6 & 7B/13B & QFormer & 8*A100 \\
        HuggingGPT & - & LLM (e.g., ChatGPT) as the core controller & - & 2023.3 & - & - & - \\
        BuboGPT & Vicuna & ImageBind, BLIP-2 & CC3M, CC12M, SBU and LAION, WaveCaps, LLaVA data & 2023.7 & 7B & QFormer & - \\
        X-InstructBLIP & Vicuna & EVA-CLIP-ViT-G/14, BEATsiter3+, ULIP-2 & DisCRn, MSRVTT, WebVid2M, MSRVTT QA & 2023.11 & 7B/13B & QFormer & 8*A100 \\
        NeXT-GPT & Vicuna-7B & ImageBind & CC3M, COCO-caption, WebVid-2M, AudioCaps, LLaVA-150K, VideoChat & 2023.9 & 7B/13B & Linear Layer & 18*A100 \\
        CoDi-2 & LLaMA & ImageBind, AudioLDM2, zeroscope v2 & MIMIC-IT, LAION-400M, AudioSet, Webvid, Instructpix2pix, AudioSet, alpaca & 2023.12 & - & Linear Layer & - \\
        SpeechGPT & LLaMA & HuBERT, HiFi-GAN & Gigaspeech, Common voice, LibriSpeech dataset, moss-002-sft-data dataset & 2023.11 & 13B & Linear Layer & 96*A100 \\
        \bottomrule
    \end{tabular}%
    }
    
    \label{tab:additional_mlm_models_audio}
\end{sidewaystable}

\subsection{Audio Tasks}
Table \ref{tab:additional_mlm_models_audio} provides a review and comparison of several representative audio MLLMs, highlighting their characteristics and challenges in terms of architecture design, multimodal fusion, dataset selection, and performance.

In terms of model architecture, most MLLMs utilize LLMs (such as Vicuna and LLaMA) as backbone networks and integrate advanced vision models (such as ImageBind and BLIP-2) and audio models (such as HuBERT and AudioLDM2) to achieve cross-modal alignment between text, image, and audio. For example, BuboGPT~\cite{zhao2023bubogpt} employs the Vicuna language model, ImageBind and BLIP-2 vision models, and the QFormer alignment module, and it is trained on multiple large-scale datasets (such as CC3M and WaveCaps), demonstrating exceptional audio understanding and generation capabilities. Conversely, SpeechGPT~\cite{zhang2023speechgpt} uses the LLaMA language model, the HuBERT speech model, and the HiFi-GAN speech synthesis model, achieving high-quality speech recognition and synthesis when trained on datasets like Gigaspeech.

Multimodal fusion technology is a critical component of audio MLLMs, responsible for mapping text, image, and audio information into a shared semantic space. Current mainstream fusion methods include simple linear layers (as used in PandaGPT~\cite{su2023pandagpt} and NeXT-GPT~\cite{wu2023next}) and attention mechanisms (as seen in Video-LLaMA~\cite{zhang2023video} and BuboGPT's QFormer structure). While attention mechanisms can capture more complex inter-modal interactions compared to linear layers, they also incur higher computational costs.

The choice and quality of datasets are crucial for the performance of audio MLLMs. Existing models are primarily trained on large-scale text-image datasets (such as CC3M and LAION), video datasets (such as WebVid-2M and MSRVTT), and audio datasets (such as AudioSet and LibriSpeech). However, these datasets have limitations in terms of annotation quality, noise levels, and domain diversity, which constrain the generalization capabilities of the models.

Although audio MLLMs have achieved impressive results in various benchmark tests, their application in real-world scenarios still faces numerous challenges. For instance, their performance may significantly degrade in noisy environments or with varying accents. Moving forward, research on audio MLLMs can focus on developing more efficient and lightweight model architectures to reduce computational costs while maintaining performance. Additionally, exploring domain adaptation and continual learning techniques can enhance model robustness and adaptability across different scenarios. Incorporating external knowledge and commonsense reasoning can improve the models' semantic understanding and generation capabilities. Furthermore, enhancing model interpretability and controllability is essential for enabling user customization and meeting specific constraints. 

\section{Discussion and Conclusion}
We provide a systematic introduction to the current state-of-the-art in MLLMs. From the task perspective, we analyze the application of MLLMs in different tasks, Including contributions and performance in natural language, and visual and auditory tasks.  However, despite their success in real-world applications, these models still face several challenges.

\textbf{Interpretability of multimodal information fusion}: Although MLLMs can significantly enhance task performance by integrating information from different modalities, their complexity and the opacity of their internal mechanisms often result in reduced interpretability. Understanding how different modalities are combined and the contribution of each modality to the final decision is crucial for a thorough analysis of multimodal data interactions. Currently, the integration of multimodal information is frequently treated as a ``black box", making it challenging for users to comprehend why and how the model reaches specific decisions. Future research should focus on addressing several key questions: How can we analyze the decision-making process in MLLMs based on different modal information? How can we enhance the trustworthiness of multimodal models in critical areas like healthcare and finance by improving their explanatory capabilities? Developing a deeper understanding of the interaction mechanisms between different modalities and their relevance to specific tasks will be essential for advancing model interpretability.

\textbf{The evolutionary direction of MLLMs}: As model sizes continue to expand, there is an ongoing debate about whether MLLMs should pursue a ``big and comprehensive" approach or a ``small and specialized" one. On one hand, creating generalized Artificial Intelligence Generated Content systems to address a wide range of real-world tasks is an attractive goal. However, increasing evidence suggests that smaller, more targeted models can deliver superior performance in specific domains. Therefore, finding the right balance between generality and specialization—ensuring that multimodal large language models can excel in specific tasks while maintaining strong generalization capabilities—has become a pivotal issue for future research.

\textbf{Security and Ethical Issues of MLLMs}: As MLLMs process vast amounts of data, they inevitably encounter security issues such as privacy leakage and data bias. Addressing these challenges is not only a technical endeavor but also a measure of researchers' social responsibility. Establishing robust ethical standards and security measures to ensure that model development and deployment adhere to societal and moral norms is a crucial issue that demands urgent and in-depth consideration.



\bibliographystyle{IEEEtran.bst}
\bibliography{mybib}

\end{document}